\documentclass{article}

\usepackage[preprint, nonatbib]{neurips_2020}

\usepackage[utf8]{inputenc} 
\usepackage[T1]{fontenc}    
\usepackage{hyperref}       
\usepackage{url}            
\usepackage{booktabs}       
\usepackage{amsfonts}       
\usepackage{nicefrac}       
\usepackage{microtype}      
\usepackage{graphicx}

\usepackage{algorithmic}
\title{Semi-supervised Learning of Galaxy Morphology using Equivariant Transformer Variational Autoencoders}

\author{%
  Mizu Nishikawa-Toomey\\
  Department of Computer Science\\
  University of Oxford\\
  \texttt{mizunt@gmail.com} \\
  \And
  Lewis Smith \\
  Department of Computer Science\\
  University of Oxford\\
  \texttt{lewis.smith@kellogg.ox.ac.uk} \\
  \AND
  Yarin Gal \\
  Department of Computer Science\\
  University of Oxford\\
  \texttt{yarin@cs.ox.ac.uk} \\
}

\begin{document}

\maketitle

\begin{abstract}
The growth in the number of galaxy images is much faster than the speed at which these galaxies can be labelled by humans. However, by leveraging the information present in the ever growing set of unlabelled images, semi-supervised learning could be an effective way of reducing the required labelling and increasing classification accuracy. We develop a Variational Autoencoder (VAE) with Equivariant Transformer layers with a classifier network from the latent space. We show that this novel architecture leads to improvements in accuracy when used for the galaxy morphology classification task on the Galaxy Zoo data set. In addition we show that pre-training the classifier network as part of the VAE using the unlabelled data leads to higher accuracy with fewer labels compared to exiting approaches. This novel VAE has the potential to automate galaxy morphology classification with reduced human labelling efforts.
\end{abstract}

\section{Introduction}
The Galaxy Zoo data set \cite{10.1111/j.1365-2966.2008.13689.x} consists of 250,000 of the brightest galaxies from the Sloan Digital Sky Survey classified by volunteers who were asked questions based on features of the galaxies such as
“Smooth, featured or artefact” or “Bar or no bar”. The data set consists of the total
number of responses for each answer to each question, and the corresponding galaxy
image. However, the number of galaxy images continues to grow at a rate that is not possible to be classified by humans. It would take 5 years to collate 40 volunteer responses for each image in the Galaxy Zoo data set at the current response rate \cite{mike}. The problem at hand is one that is widely prevalent in the modern age; how
can the abundant amounts of unlabelled data be used to compliment the learning task
using the limited labelled data. 
Prior research conducted on the data set \cite{mike} uses active learning \cite{Houlsby2011BayesianAL}
to select the most informative galaxies for classification by volunteers, showing
improvements in the accuracy of the classification task using this technique. However, as far as we know
any form of semi-supervised learning where information from the unlabelled images
themselves are used to train a neural network for the classification task has yet
to be explored for this problem.

We develop a novel VAE \cite{Kingma2014AutoEncodingVB} architecture for semi-supervised learning and compare the performance of fully-supervised learning using the same amount of labelled data. 
VAEs with a classifier network from the latent variables have previously been shown to be beneficial for semi-supervised image classification tasks \cite{ssvae}. The novelty of our method comes from the introduction of Equivariant Transformer layers \cite{eqt} in the VAE, and the classifier weights being back-propagated along with the weights of the encoder. The Equivariant Transformer layers remove the dependence of pose from the latent representation of the galaxy, and in turn removes the dependence of pose on the classification, improving on data efficiency. This novel architecture for a VAE with Equivariant Transformer layers suits image data sets such as galaxy images, where the classification is independent of the pose of the image. 
In addition to the generative model p(x|z), VAEs learn the approximate posterior
distribution over the latents q(z|x). This opens up many avenues for architectures
for semi-supervised learning compared to other generative models such as Generative Adversarial Networks (GANs) \cite{gan} that only model the generative distribution. In addition, using this probabilistic framework instead of a standard Autoencoder (AE), allows for the possibility for other downstream tasks such as active learning using the posterior over the latents. 

We show that this new architecture for semi-supervised learning results in enhanced performance
of the classification task for this problem compared to existing semi-supervised or supervised learning techniques.

\section{Methods}
\subsection{Probabilistic framing of the problem}
Each galaxy image is shown to multiple volunteers in the Galaxy Zoo data set. The data set consists of the number of positive responses to each answer for each question for a galaxy image. The number of positive responses to a particular question is modelled as a multinomial distribution parameterised by a vector $k$ as done in previous work by Walmsley et al. \cite{mike}. The length of the vector $k$ is the number of possible answers to the question, and each element corresponds to the probability of a positive response for a particular answer. By using the negative log likelihood of this multinomial distribution as the objective function, the neural network will learn to predict the vector $k$ for a particular question for each galaxy image using the maximum likelihood estimate.
\begin{equation}
    k = \textrm{NeuralNetwork}_{\theta}(x)
\end{equation}
\begin{equation}
    p(y) = \textrm{Multinomial}(k)
\end{equation}
In these experiments, the question that was investigated was “Smooth, featured or artefact”.

\subsection{Model architecture}
The novel architecture consisted of a VAE \cite{Kingma2014AutoEncodingVB} with Equivariant Transformer layers \cite{eqt} where a canonical pose is predicted in the encoder, and an inverse transformation from the canonical pose to the original coordinate system is performed in the decoder. A two-layer neural network is used to classify images from the latent variables which are disentangled from the pose of the image. The cosmological principle which states that the universe is homogeneous and isotropic are grounds to base the assumption that the morphology classification of galaxies are independent of the pose of the galaxy.

Equivariant Transformers allow the 
network to learn a canonical rotation for the galaxy image that is beneficial to minimising the objective function. This results in removing the dependence of the pose on the latent variable, reducing redundancy in the data set and increasing data efficiency. In turn, the classification of the galaxies will also be independent of the pose.  
Equivariant Transformers predict the pose parameters of transformations whilst maintaining self-consistency.  This is done by transforming the image in to its canonical coordinate system by the transformation $\rho$ which satisfies:

\begin{equation}
    \rho(T_{\theta}x) = \rho(x) + \sum_{i=1}^{k} \theta_{i}e_{k} 
\end{equation}

where $\theta_{i}$ are the pose parameters and $e_{k}$ are the basis vectors. This results in a transformation applied to the image in the original coordinate system becoming a translation in the canonical coordinate system. A rotation by $\theta$ for example, corresponds to a translation by $\theta$ in the angular coordinate in polar coordinates, making polar coordinate systems an example of a canonical co-ordinate system with respect to rotation.
After the image is transformed in to its canonical coordinates, a function which is invariant to translation, such as a convolutional neural network is applied to predict the pose parameters of the image in its canonical coordinates. 

The weights of the classifier were back-propagated through the weights of the encoder when minimising the objective function for the classifier. This differs from an already established method of classification from the latent space of a VAE where only the weights of the classifier are adjusted with respect to the objective function of the classification task, but not the weights of the encoder, described in \cite{ssvae} as the M1 model. Performance improvements were observed for this new method of updating the encoder weights as well as the classifier weights.
The VAE is trained in the semi-supervised experiments using a surrogate loss, the Evidence lower bound, or the \textsc{elbo}, given by:

\begin{equation}
 \mathcal{L}(x)= \mathbb{E}_{q_{\phi}(z|x)} [\log \, p_{\theta}(z,x)] - \mathbb{E}_{q_{\phi}(z|x)} [\log \, q_{\phi}(z|x)] 
\end{equation}

The \textsc{elbo} provides a lower bound for the intractable marginal likelihood $p(x)$. Minimising the negative of the \textsc{elbo} is equivalent to minimising the KL divergence of the approximate posterior and the real posterior. In training VAEs, the negative \textsc{elbo} is minimised with respect to $\theta$ which parameterises the generative model, and $\phi$ which parameterises the variational distribution. 

\subsection{Comparison of supervised and un-supervised learning}
\begin{figure}[h]
    \centering
    \includegraphics[width=0.6\textwidth]{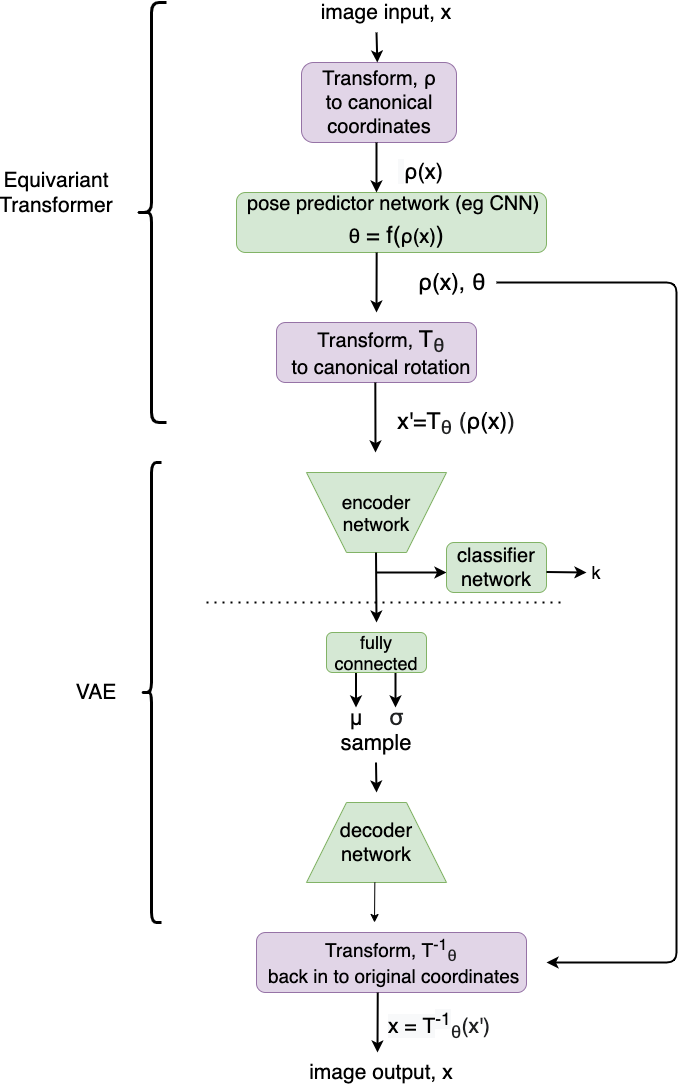}
        \caption{The semi-supervised training regime using pose prediction. The fully-supervised training regime is denoted by the network above the dotted line.}
    \label{fig:pose}  
\end{figure}

To compare the results of the semi-supervised architecture to an equivalent supervised architecture, the following setup was used. In the fully-supervised training regime, the decoder of the VAE was neglected, and the classifier and the encoder was back-propagated through, using only labelled data. There is only one objective function to be minimised, which is the negative log likelihood of the prediction. In the semi-supervised training regime, there are two objective functions to minimise, the negative \textsc{elbo} for training the VAE as well as the negative log likelihood of the classifier. This setup can be seen in Figure \ref{fig:pose}.

Alternating steps were taken in the semi-supervised training regime between training the VAE with unlabelled data and training the classifier and the encoder with labelled data.

\section{Results}

Experiments were conducted with varying amounts of unlabelled data, starting from 100 to 1200 labelled galaxy images. The results are shown in Table \ref{table1}.

Semi-supervised learning outperforms fully-supervised learning of galaxy morphology classification for the number of labelled images tested in these experiments. 
To investigate the further use cases of the semi-supervised architecture, experiments were conducted where the VAE was trained prior to the classifier in an unsupervised manner, then the classifier and the encoder was trained using the pre-trained weights of the encoder.

\begin{table}[h!]
  \caption{Root mean squared error for semi-supervised and supervised learning using differing amounts of labelled data. The performance of two-step training where the VAE is trained prior to the classifier compared to when then the gradient steps of the VAE and classifier are alternated.}
  \label{table1}
  \centering
  \begin{tabular}{lllll}
    \toprule

    Number of labelled images & 100 & 300   & 800  & 1200\\
    \midrule
    Fully supervised  & 0.56 & 0.31   & 0.25  & 0.24\\
    Semi-supervised 2-step training of VAE and classifier  & 0.37 & 0.28  & 0.25  & 0.25\\
    Semi-supervised, alternating steps of VAE and classifier  & 0.35 & 0.24  & 0.20  & 0.21    \\

    \bottomrule
  \end{tabular}
\end{table}

It was evident that unsupervised pre-training of the weights of the encoder as part of a VAE was beneficial to the classification task, compared to random initialisation of the weights when using small amounts of labelled data. The performance gains start to diminish after 800 labelled images, this is thought to be as a result of the information stored in the pre-trained weights becoming erased as more gradient steps are taken to optimise the network for the classification objective only.
One of the benefits of a two-step training regime is that training the VAE on the unlabelled data could be done prior to obtaining labelled data on the images. Once labels are available, the weights of classifier and encoder can be fine-tuned for the classification task, and the performance benefits of semi-supervised learning can be seen immediately without having to go through the computationally expensive process of training the VAE on the vast amounts of unlabelled data at the same time as training the classifier using labelled data.

\section{Conclusion}
Semi-supervised learning of galaxy morphology classification using a VAE \cite{Kingma2014AutoEncodingVB} resulted in enhanced performance of classification compared to fully-supervised learning. The novel semi-supervised learning architecture consisted of a VAE with Equivariant Transformers \cite{eqt} in the encoder, where the classifier was back-propagated along with the weights of the encoder. Further experiments showed that initialising the classifier by using an encoder that was trained as part of a VAE resulted in enhanced performance of the classifier than when the classifier weights were randomly initialised. This removed the overhead of training the VAE alongside training the classifier when performing semi-supervised training whilst still retaining some of the performance gains from semi-supervised learning. This novel method of semi-supervised learning of galaxy morphology classification has applications in scenarios where labelled data is scarce. 
\clearpage

\section{Broader impact}
Researchers who do not have access to fully labelled data sets would benefit from this work. They would be able to leverage the information stored in unlabelled data to achieve performance gains in their classification task. This means less resources can be spent on tasking individuals to label data sets. For example, if the methods introduced were to be used to classify cancerous or benign skin cancer tumours, expert dermatologists would have to spend many hours labelling images of cancerous and benign tumours, at the expense of other valuable work that they could pursue.  

On the other-hand, the use of these methods could result in researchers being dependent on fewer labelled data as a result of increased accuracy, and therefore the algorithm will be more reliant on the scarce number of labelled images. If the number of labelled images is too few, over fitting of the algorithm would lead to some edge cases being in-correctly classified by the algorithm. Care must be taken that the algorithm does not over-fit the data, when working with small sets of labelled data. 

\bibliographystyle{unsrt}
\bibliography{Bibliography}

\end{document}